\documentclass[letterpaper]{article} % DO NOT CHANGE THIS
\usepackage{aaai2026}  % DO NOT CHANGE THIS
\usepackage{times}  % DO NOT CHANGE THIS
\usepackage{helvet}  % DO NOT CHANGE THIS
\usepackage{courier}  % DO NOT CHANGE THIS
\usepackage[hyphens]{url}  % DO NOT CHANGE THIS
\usepackage{graphicx} % DO NOT CHANGE THIS
\usepackage{natbib}  % DO NOT CHANGE THIS AND DO NOT ADD ANY OPTIONS TO IT
\usepackage{caption} % DO NOT CHANGE THIS AND DO NOT ADD ANY OPTIONS TO IT
\usepackage{multirow}
\usepackage{booktabs}

\usepackage{algorithm}
\usepackage{algorithmic}

\usepackage{newfloat}
\usepackage{listings}
\DeclareCaptionStyle{ruled}{labelfont=normalfont,labelsep=colon,strut=off} % DO NOT CHANGE THIS
\floatstyle{ruled}
\newfloat{listing}{tb}{lst}{}
\floatname{listing}{Listing}
\title{HierOctFusion: Multi-scale Octree-based 3D Shape Generation via Part-Whole-Hierarchy Message Passing}
\author{
    Xinjie Gao\textsuperscript{\rm 1}\equalcontrib,
    Bi'an Du\textsuperscript{\rm 1}\equalcontrib,
    Wei Hu\textsuperscript{\rm 1}
}
\affiliations{
    \textsuperscript{\rm 1}Wangxuan Institute of Computer Technology, Peking University\\
    gxj1914779162@stu.pku.edu.cn, pkudba@stu.pku.edu.cn, forhuwei@pku.edu.cn
}

\usepackage{bibentry}
\begin{document}

\maketitle

\begin{abstract}
\vspace{-0.05in}
3D content generation remains a fundamental yet challenging task due to the inherent structural complexity of 3D data. While recent octree-based diffusion models offer a promising balance between efficiency and quality through hierarchical generation, they often overlook two key insights: 1) existing methods typically model 3D objects as holistic entities, ignoring their semantic part hierarchies and limiting generalization; and 2) holistic high-resolution modeling is computationally expensive, whereas real-world objects are inherently sparse and hierarchical, making them well-suited for layered generation. Motivated by these observations, we propose HierOctFusion, a part‑aware multi‑scale octree diffusion model that enhances hierarchical feature interaction for generating fine‑grained and sparse object structures. Furthermore, we introduce a cross‑attention conditioning mechanism that injects part‑level information into the generation process, enabling semantic features to propagate effectively across hierarchical levels from parts to the whole. Additionally, we construct a 3D dataset with part category annotations using a pre-trained segmentation model to facilitate training and evaluation. Experiments demonstrate that HierOctFusion achieves superior shape quality and efficiency compared to prior methods.
\end{abstract}

%\begin{links}
%    \link{Code}{https://aaai.org/example/code}
%    \link{Datasets}{https://aaai.org/example/datasets}
%    \link{Extended version}{https://aaai.org/example/extended-version}
%\end{links}

\vspace{-0.05in}
\section{Introduction}
\vspace{-0.05in}
Content generation is a fundamental task in computer science. It involves training neural networks to automatically generate text, images, audio or video, holding significant application value and research potential in numerous real-world scenarios. Despite significant advances in Variational Autoencoders (VAEs)~\cite{kingma2013VAE}, Generative Adversarial Networks (GANs)~\cite{goodfellow2014GAN}, and diffusion models~\cite{ho2020DDPM, song2022DDIM, rombach2022SD}, generating 3D content remains difficult. Unlike 2D generation, 3D objects and scenes introduce substantially higher complexity, making it challenging to achieve both high fidelity and computational efficiency.

Current 3D generation methods fall into two categories. The first adapts pre-trained 2D generative models~\cite{poole2022dreamfusion, lin2023magic3d}, leveraging their strong image priors but relying on expensive multi‑view optimization that is both slow and prone to inconsistency. The second trains 3D generators directly on object datasets~\cite{hui2022neural, gupta20233dgen, chou2023diffusionsdf, li2023diffusion, cheng2023sdfusion}, eliminating multi‑view loops at the cost of being highly sensitive to the chosen 3D representation.

\begin{figure*}[!t]
\centering
\includegraphics[width=0.95\textwidth]{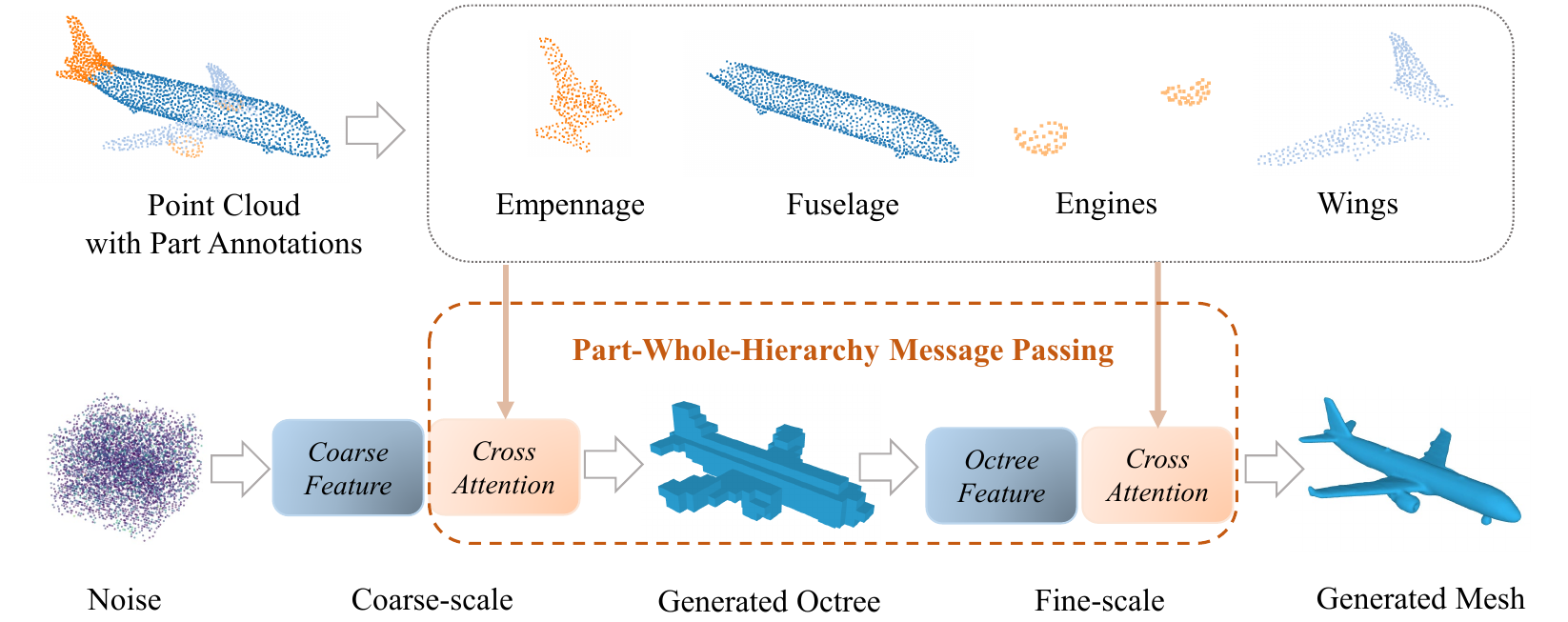} % Reduce the figure size so that it is slightly narrower than the column.
\caption{HierOctFusion is a part‑aware multi‑scale octree diffusion model that enhances hierarchical feature interaction via a cross‑attention conditioning mechanism. It is capable of generating high-quality 3D objects in various scenarios.}
\label{fig1}
\end{figure*}

The quality and efficiency of 3D generation depend on two interrelated challenges: choosing a representation that can capture complex geometry inexpensively, and designing a generation process that is appropriate for that representation. Prior works~\cite{wang2022dual, xiong2024octfusion} address this by encoding shapes as volumetric octrees and applying a diffusion model that incrementally deepens the tree. This strategy preserves the smoothness of implicit features while exploiting the explicit hierarchy of spatial partitions. However, existing octree‑based approaches treat every cell identically, distinguishing regions only by tree depth and overlooking the natural part‑whole structure of 3D objects. As a result, they struggle to generalize across varied shapes and miss opportunities for targeted refinement. In contrast, 3D part assembly inherently follows a local‑to‑global paradigm, where individual semantic parts are constructed and then combined. Building on the part‑whole hierarchy mechanism~\cite{du2024generative}, we propose explicitly injecting part‑level cues into the octree diffusion process, enabling the model to focus on semantically meaningful regions and produce coherent and high‑quality shapes.

In this paper, we introduce HierOctFusion, a multi‑scale octree diffusion framework that incorporates part‑whole hierarchy through message passing. Extending the octree‑based diffusion of~\cite{xiong2024octfusion}, our model injects local part‑level cues as the octree deepens, directing the model’s capacity towards semantically meaningful regions and improving generation fidelity. We represent each object as a voxelized octree built via recursive partitioning of densely sampled point clouds, encode each leaf node with a pre-trained VAE~\cite{kingma2013VAE}, and decode the refined latents into a signed distance field. The resulting SDF is converted to a mesh via Marching Cubes. By combining explicit hierarchical structure with learned latent features, HierOctFusion achieves higher‑quality 3D shape synthesis without sacrificing efficiency.

Specifically, we propose a multi-scale octree diffusion model that fully leverages local semantic part information. In the coarse-scale, the network performs coarse‑grained generation: it learns to denoise inputs and produce a shallow‑depth octree, capturing the object’s global structure and overall shape distribution. The fine-scale focuses on fine-grained refinement. Taking the output of the coarse-scale as input, this phase further optimizes the octree depth by depth, generating a deeper octree and subsequently converting it into a mesh. Inspired by Stable Diffusion~\cite{rombach2022SD}, we innovatively introduce the part feature as a conditional input to the diffusion model. The part feature extracted by a pre-trained encoder is treated as semantic conditions and incorporated into the network through a hierarchical cross-attention-based mechanism. During optimization at different octree depths, the part feature interacts with the octree feature via cross-attention, enabling part-whole-hierarchy message passing and precise control over local details.

To verify the effectiveness of our method, we conducted a comprehensive evaluation on the widely used ShapeNet dataset~\cite{shapenet2015} for the 3D single object generation task. Given the small amount of data in ShapeNetPart~\cite{yi2016scalable}, we used a pre-trained segmentation model to build a 3D dataset ShapeNet-Seg with part category annotations to make up for this shortcoming. Experimental results demonstrate that our approach achieves superior generation quality compared to existing methods while maintaining generation efficiency comparable to the original octree-based diffusion model. Notably, our framework delivers improved fidelity on objects with intricate structural details. The main contributions of this paper are summarized as follows.
\begin{itemize}
    \item We introduce HierOctFusion, a diffusion model that integrates part‑level priors into a multi‑scale octree hierarchy, enabling accurate generation of fine‑grained and sparse 3D structures.
    \item We develop a hierarchical cross‑attention mechanism that injects semantic part features at each octree depth, facilitating effective message passing from parts to the whole.
    \item We construct a dataset ShapeNet‑Seg by applying a pret-rained 3D segmentation model to ShapeNet, providing part category annotations for rigorous training and evaluation. Our experiments show that HierOctFusion outperforms existing methods in quality and efficiency.
\end{itemize}

\vspace{-0.05in}
\section{Related Work}
\vspace{-0.05in}
\subsection{3D Object Representation}
3D object representation is a fundamental research problem in the fields of computer graphics and 3D vision. Selecting an appropriate 3D representation method is the primary task in 3D research, directly influencing the efficiency and quality of subsequent processing. Commonly used representation methods include voxel~\cite{maturana2015voxnet, schwarz2022voxgraf, sella2023vox}, point cloud~\cite{li2018pointcnn, guo2020deep, du2021self, chen2022deep}, mesh~\cite{kato2018neural, wang2018pixel2mesh, groueix2018papier}, SDF~\cite{park2019deepsdf, mu2021sdf, mittal2022autosdf}, etc. In recent years, NeRF~\cite{mildenhall2021nerf, shue20233d, zhang20233dshape2vecset} and 3D-GS~\cite{kerbl20233dgs, du2024auggs, li2025geometry} have also achieved widespread development and application. Recently, some works~\cite{wang2022dual, xiong2024octfusion} employ the octree as the representation method. This representation combines the benefits of both implicit neural representations and explicit spatial structures. DualOctreeGNN~\cite{wang2022dual} uses an octree to partition the 3D volume into adaptive voxels and learns a neural MPU to generate continuous volumetric fields. OctFusion~\cite{xiong2024octfusion} converts the point cloud to an octree and all leaf nodes of the octree are associated with a latent feature, which is decoded to SDF by a MLP. Our representation method largely follows OctFusion, with the additional incorporation of part-level information.

\subsection{3D Object Generation}
3D part assembly and object generation have each drawn considerable interest in recent 3D vision research. In 3D part assembly, the goal is to determine the positions and orientations of a set of unlabelled components so that their shapes and surfaces align to form a unified object~\cite{zhan2020generative, li2020learning, xu2023unsupervised}. The Part Whole Hierarchy framework~\cite{du2024generative} addresses this by using a message passing network to aggregate individual components into progressively larger clusters, thereby capturing geometric relationships at multiple scales. In the domain of 3D generation, researchers have extended variational autoencoders~\cite{kingma2013VAE}, generative adversarial networks~\cite{goodfellow2014GAN} and diffusion models~\cite{ho2020DDPM, song2022DDIM, rombach2022SD, du2024multi} from images to volumetric data. These volumetric methods often require very large memory and compute budgets in order to represent fine detail. To overcome this, we follow the octree diffusion approach of OctFusion~\cite{xiong2024octfusion}. In our method, the network is trained in two scales with an octree whose subdivision depth increases gradually during generation. By aligning the progressive refinement of the octree with our part-based hierarchy module, we produce detailed and coherent 3D objects while avoiding excessive resource consumption.

\vspace{-0.05in}
\section{Method}
\vspace{-0.05in}
\subsection{Overview}
\begin{figure*}[t]
\centering
\includegraphics[width=0.95\textwidth]{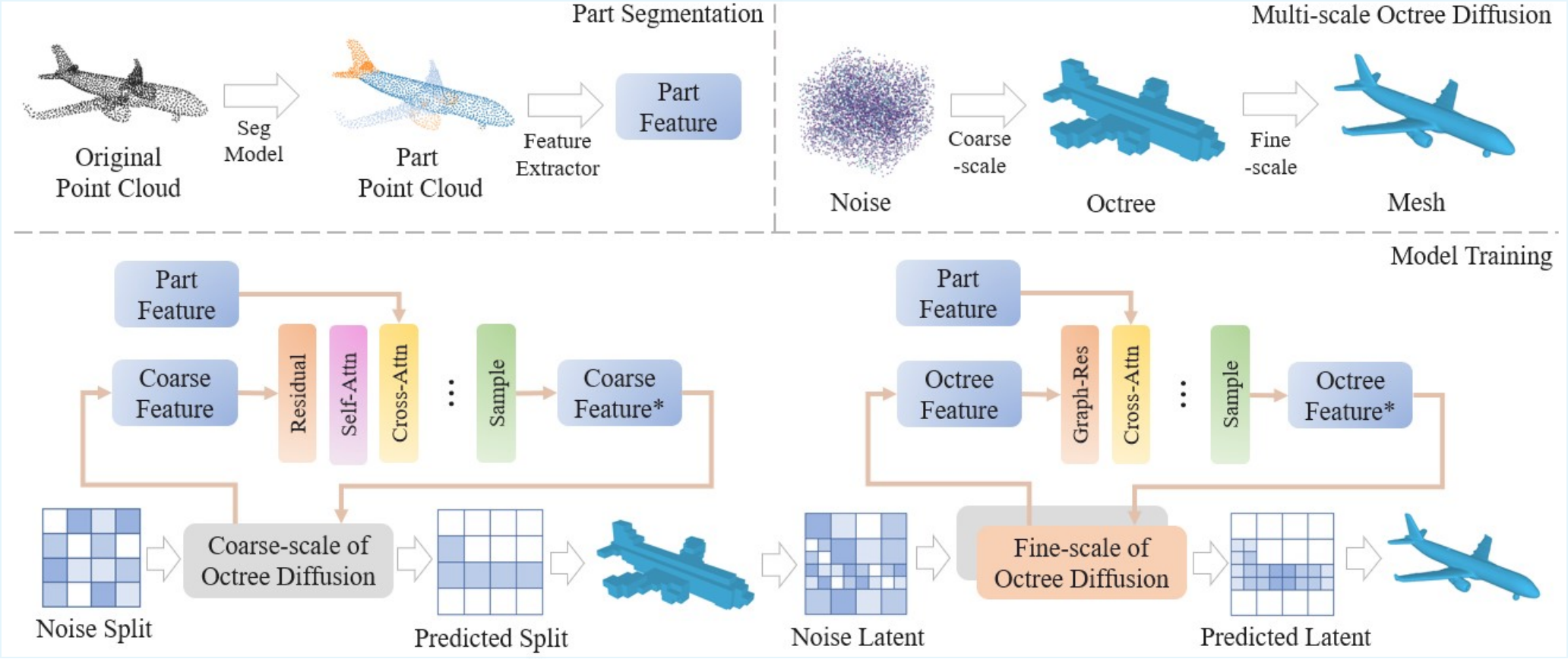} % Reduce the figure size so that it is slightly narrower than the column.
\caption{Overview of HierOctFusion. Top: The input point cloud is decomposed into multiple semantic parts. A feature encoder then extracts part features from these segmented components. Bottom: The generative model training consists of two scales. In the coarse-scale, the noisy input is denoised to reconstruct a low-depth octree. In the fine-scale, the previously generated octree undergoes refinement to extend its depth. Part features are incorporated into the model via a cross-attention mechanism.}
\label{fig2}
\end{figure*}

Our goal is to integrate the local information of semantic parts with the hierarchical generation process of octrees so that part-level features flow naturally through the octree’s varying depths. To realize this, we introduce a part–whole–hierarchy diffusion model built on an octree backbone. In our framework, part-level features are integrated into the diffusion process through a cross-attention conditioning mechanism, which guides the octree’s depth-wise refinement. Figure \ref{fig2} provides an overview of our pipeline. We begin by constructing the ShapeNet-Seg dataset with rich part annotations. We then detail our part–whole–hierarchy cross-attention module and demonstrate how it enhances a multi-scale octree diffusion model to support part-aware generation.

\subsection{ShapeNet‑Seg Dataset Construction}
Semantic part information for 3D objects depends on datasets annotated at the part level, yet most existing repositories do not provide such detail. For example, ShapeNetCore~\cite{shapenet2015} comprises 51,300 high-quality CAD models across 55 common object categories, yet it does not provide any part-level annotations. Its companion dataset, ShapeNetPart~\cite{yi2016scalable}, offers part-level segmentation but covers only 16 categories and 16,880 models. To bridge this gap, we have created and released ShapeNet-Seg, which extends ShapeNet with comprehensive part annotations. In the following section, we first introduce the part segmentation model we selected and then detail the preprocessing pipeline that generates these annotations. Our main objective is to maintain geometric consistency in the segmentation outputs and ensure they integrate smoothly with downstream generation tasks, thereby making ShapeNet-Seg an effective resource for octree-based synthesis.

\subsubsection{Part segmentation model.}
To build ShapeNet-Seg, we first extract a subset of 3D models from the ShapeNet~\cite{shapenet2015} dataset, including airplanes, cars, chairs, and tables, to evaluate pre-trained segmentation models. We compare various architectures for inference speed and ease of integration, ultimately selecting DGCNN~\cite{phan2018dgcnn} as our segmentation backbone. By running a fine-tuned DGCNN on 3D models from ShapeNet, we obtain consistent and high-quality part labels. These precise annotations form the core of ShapeNet-Seg and provide reliable semantic cues for our subsequent octree-based generation pipeline.

\subsubsection{Data preprocess.}
We preprocess 3D models from four categories in ShapeNet (airplane, car, chair and table) by converting each mesh into a high density point cloud of approximately 200 000 points. This sampling density preserves geometric detail for subsequent octree construction and limits information loss during part segmentation. Each dense point cloud is uniformly downsampled to 2048 points and processed by a fine-tuned DGCNN to generate semantic part annotations. Based on the structural characteristics of each class we divide airplanes, cars and chairs into four parts and tables into three parts. The resulting part category labels are embedded as additional attributes in the point clouds to produce the ShapeNet Seg dataset, a richly annotated point cloud collection that provides a robust foundation for our octree-based 3D object generation pipeline.

\subsection{Part‑Whole‑Hierarchy Cross‑Attention Module}
Following the creation of ShapeNet-Seg with part-level annotations, the next objective is to propagate semantic part information through the octree’s hierarchical structure. To this end, we propose a part–whole–hierarchy cross-attention module. This section proceeds as follows. First, the part–whole–hierarchy mechanism is formalized. Then, the detailed architecture of the cross-attention module is presented.

\subsubsection{Part-whole hierarchy message passing.}
The concept of a part-whole hierarchy is realized in \cite{du2024generative} by introducing super-parts that link individual components to the complete object in three tiers. We adapt this mechanism for 3D generation by segmenting each instance into semantic parts and injecting their features into successive octree-based diffusion refinements. More specifically, in a manner analogous to conditional control, we inject part-aware features into the generation process at each octree depth level via a cross-attention mechanism. Our approach enables the model to effectively leverage the localized structural information encapsulated within the semantic parts to refine the generation of corresponding regions. This global-to-local message passing markedly improves the model’s capacity to generate intricate geometry.

\subsubsection{Module design.}
Cross-attention, a core component of the Transformer architecture~\cite{vaswani2017attention}, excels at fusing heterogeneous features. We introduce a lightweight module built on this mechanism. First, octree features produced during our pipeline pass through convolutional layers to form the Query $Q$. Separately, part features from the encoder undergo two convolutional transformations to generate the Key $K$ and Value $V$. We compute the similarity matrix by taking the dot product of $Q$ and $K$, then scale and normalize the result to obtain the attention weights. Conceptually, this operation uses Query to retrieve relevant Key elements, with each attention weight indicating the matching score of a specific Query–Key pair. Larger matching scores yield higher attention weights. Next, we apply those weights to Value via a weighted sum, producing the updated feature representation. This process can be formally expressed as:
\begin{equation}
\label{eq1}
    L_{updated} = softmax(\frac{QK^T}{\sqrt{d_k}})V,
\end{equation}
where $L_{updated}$ is the updated octree feature, and $d_k$ is the dimension of Query and Key. In our design, each octree node’s features serve as Query to perform retrieval within the part feature space. The resulting attention weights guide the update of octree features at each node. In this way, our module achieves effective fusion of object-level and part-level representations.

\subsection{Part‑Aware Multi‑Scale Octree Diffusion}
After building the part–whole-hierarchy cross-attention module, we integrate it into the diffusion framework to enable part-aware, multi-scale octree diffusion. First, we give a concise overview of diffusion models; then we present a detailed description of our modified octree diffusion architecture.

\subsubsection{Diffusion models.}
Diffusion models are a widely adopted class of generative models that operate by sampling noise from a simple distribution and then iteratively denoising it to generate target outputs. These models consist of two key phases: the forward process and the reverse process. The forward process is a Markov chain that gradually adds Gaussian noise to the data according to a predefined schedule over discrete timesteps. This process can be formally expressed as:
\begin{equation}
\label{eq2}
    q(x_t|x_{t-1})=\mathcal{N}(x_t;\sqrt{1-\beta_t}x_{t-1},\beta\mathbf{I}),
\end{equation}
where $x_t$ represents the octree data at timestep $t$, $\beta_t$ denotes the noise schedule hyperparameters controlling the variance of the Gaussian noise.If we pre-define $\alpha_t=\prod_{i=0}^t(1-\beta_i)$, mathematical derivation from Eq. \ref{eq2} yields the closed-form expression:
\begin{equation}
\label{eq3}
    x_t=\sqrt{\alpha_t}x_0+\sqrt{1-\alpha_t}\epsilon,
\end{equation}
where $\epsilon$ represents standard Gaussian noise. Since $\beta_t$ is a hyperparameter, $\alpha_t$ can be precomputed in advance, thereby allowing the determination of $x_t$ at any time step to facilitate the subsequent calculation of the loss function.The reverse process is a stepwise denoising procedure that begins by sampling noise from $\mathcal{N}(0,\mathbf{I})$ and progressively refines it to restore the original data distribution, thereby achieving the generative process. Since directly estimating $q(x_{t-1}|x_t)$ is intractable, a neural network $\mathcal{F}(x_t,t)$ can be employed to model the denoising transition from $x_t$ to $x_0$. This network is trained using the following loss function:
\begin{equation}
\label{eq4}
    L_{diffusion}=\rm{E}_{\epsilon,t}||\mathcal{F}(x_t,t)-x_0||_2^2.
\end{equation}
A standard diffusion model typically consists of two distinct phases, training and sampling. Once the training process is completed using the aforementioned loss function, the sampling phase involves drawing noise from a standard Gaussian distribution and iteratively applying denoising operations to produce the final generated output.

\subsubsection{Multi-scale octree U-Net architecture.}
We utilize a single U-Net architecture that processes octree representations at multiple depths and divide its training into two sequential scales. In the coarse-scale, denoted $\mathcal{F}_1$, the network receives noisy octrees of depth four and gradually denoises them until they reach depth six. Once $\mathcal{F}_1$ training concludes, its parameters are frozen and the fine-scale, $\mathcal{F}_2$, commences. In $\mathcal{F}_2$, we add noise to depth-six octrees and then perform denoising to extend them to depth eight. We deliberately cap the maximum depth at eight to balance computational efficiency with generation fidelity. Throughout both scales, we augment the diffusion model with our cross-attention module. In $\mathcal{F}_1$, the backbone closely follows conventional 2D image diffusion U-Nets, incorporating residual blocks, self-attention layers, downsampling, and upsampling. We insert our module immediately after each self-attention layer, adapting its architecture to incorporate part-level cues. In the fine-scale, $\mathcal{F}_2$, we further exploit local information from semantic parts by placing the cross-attention mechanism after every graph residual block, adjusting its inputs accordingly. This design ensures that part features are deeply embedded throughout the denoising process, yielding a part-aware octree diffusion model capable of fully leveraging part-level information. Our multi-scale octree diffusion process can be expressed as:
\begin{equation}
\label{eq5}
    x_0 = \mathcal{F}_2(\mathcal{F}_1(x_T;d_1);d_2),
\end{equation}
where $d_1$ represents the maximum depth of the coarse-scale, and $d_2$ represents that of the fine-scale.

To produce the final 3D outputs, our framework draws part features from any exemplar object within the same category and uses these features to guide the diffusion process. After synthesizing the octree, we convert it into both a signed distance field representation and a mesh model to achieve high-quality visualization. By combining part-whole-hierarchy message passing with depth-progressive octree optimization, our approach realizes a comprehensive octree-based generative diffusion pipeline that enhances geometric detail without sacrificing efficiency.

\vspace{-0.05in}
\section{Experiments}
\vspace{-0.05in}
In this section, we present the dataset specifications in our method, followed by comprehensive comparisons with state-of-the-art approaches to validate the effectiveness of our proposed method.

\subsection{Dataset}
The primary dataset in this study is ShapeNet~\cite{shapenet2015}, which contains approximately 50 000 3D meshes spanning 55 common object categories. Following OctFusion~\cite{xiong2024octfusion}, we focus on four classes, namely airplane, car, chair and table, which include about 4 000, 7 500, 6 800 and 8 400 models respectively. To construct ShapeNet-Seg~\cite{yi2016scalable}, each mesh is first sampled to produce a dense point cloud of 200 000 points; farthest point sampling then reduces this to 2 048 points while preserving geometric coverage. Those points are passed through a pre-trained part segmentation model, which assigns four part labels to each airplane, car and chair model and three part labels to each table model. We save the 3D coordinates and part labels of all sampled points to form our part-level annotated point-cloud dataset. During diffusion training, we reuse the DGCNN encoder to extract part-level feature descriptors. These descriptors are injected into our octree-based diffusion model via the cross-attention module, guiding generation with detailed part-aware information.

\begin{table}[h]
\centering
\resizebox{.95\columnwidth}{!}{
\begin{tabular}{c cccc}
\toprule
Dataset & Number & Category & Part Annotation\\
\midrule
ShapeNetCore & 51,300 & 55 & No\\
ShapeNetPart & 17,000 & 16 & No\\
ShapeNet-Seg & 31,000 & 16 & Yes\\
\bottomrule
\end{tabular}
}
\caption{Comparison of ShapeNetCore, ShapeNetPart and ShapeNet-Seg. Number denotes the total count of models, category indicates the total number of object classes, and part annotation denotes whether the dataset contains part-level annotations.}
\label{tab1}
\end{table}

\begin{figure*}[ht]
\centering
\includegraphics[width=1.0\textwidth]{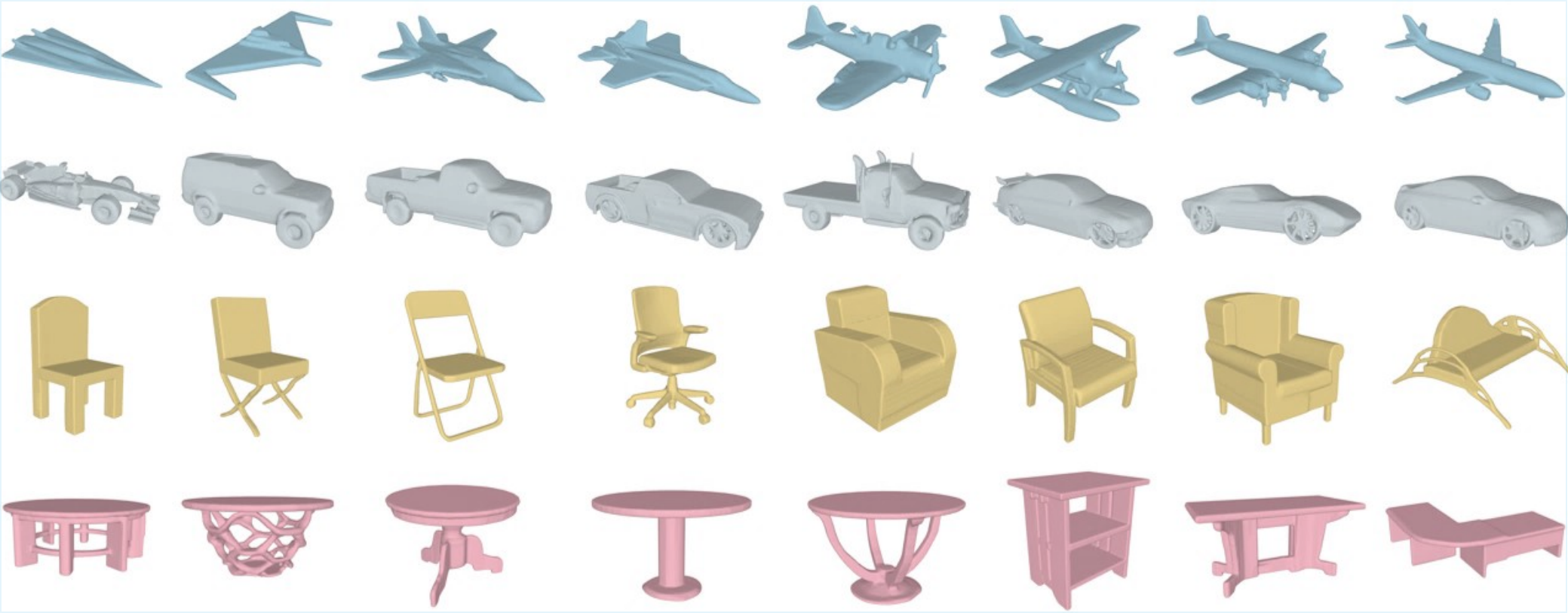} % Reduce the figure size so that it is slightly narrower than the column.
\caption{Generative results of HierOctFusion. Our method is capable of generating high-quality and diverse 3D shapes.}
\label{fig3}
\end{figure*}

\subsection{Results}

\subsubsection{Evaluation metrics.}
Consistent with previous studies \cite{zheng2023locally, zhang20233dshape2vecset, xiong2024octfusion}, we adopt the shading-image–based FID as our primary evaluation metric for generative quality. The FID score is widely used to measure how closely a set of generated images matches the distribution of real images, capturing both visual fidelity and diversity. A lower FID indicates that the generated shading images share more similar high–level characteristics with the real images, reflecting improved shape consistency, texture coherence and overall realism.

\subsubsection{Comparisons.}
We conduct comparisons with state-of-the-art approaches to validate the effectiveness of our method, including IM-GAN~\cite{chen2019learning}, SDF-StyleGAN~\cite{zheng2022sdf}, Wavelet-Diffusion~\cite{hui2022neural}, LAS-Diffusion~\cite{zheng2023locally} and OctFusion~\cite{xiong2024octfusion}.

\begin{table}[h]
\centering
\resizebox{.95\columnwidth}{!}{
\begin{tabular}{c cccc}
\toprule
Method & Airplane & Car & Chair & Table\\
\midrule
IM-GAN & 74.57 & 141.2 & 63.42 & 51.70\\
SDF-StyleGAN & 65.77 &  97.99 & 36.48 & 39.03\\
Wavelet-Diffusion & 35.05 &  N/A & 28.64 & 30.27\\
LAS-Diffusion & 32.71 &  80.55 & 20.45 & 17.25\\
OctFusion & 24.29 &  78.00 & 16.15 & 17.19\\
\midrule
HierOctFusion & \textbf{23.84} & \textbf{77.13} & \textbf{16.07} & \textbf{17.09}\\
\bottomrule
\end{tabular}
}
\caption{Comparison of our method with existing approaches using the shading-image-based FID metric. Our method achieves superior performance across all four categories—airplane, car, chair and table—outperforming existing baselines.}
\label{tab2}
\end{table}

Table \ref{tab2} reports the FID scores achieved by our approach and several baseline methods. Across all four categories, our model consistently yields lower FID values, indicating closer alignment with the real shading images. The most dramatic gains appear in the airplane and car categories. These classes feature highly complex local geometry, such as wing flaps, landing gear structures and vehicle surface embellishments, and our part-aware, multi-scale octree diffusion excels at capturing and reproducing these fine-grained details. In contrast, the chair and table categories involve comparatively simpler shapes and fewer intricate features. Existing generative techniques already attain reasonably low FID scores, so the absolute improvements achieved by our method are smaller. Nevertheless, our model still matches or slightly betters the state-of-the-art for chairs and tables, demonstrating its broad applicability. Figure \ref{fig3} provides qualitative examples of the shading images synthesized by HierOctFusion after training separate models for each category on the ShapeNet-Seg dataset. In these visualizations, one can observe that our method generates crisp surface shading and preserves semantic part boundaries, further corroborating the results.

\begin{figure}[h]
\centering
\includegraphics[width=0.9\columnwidth]{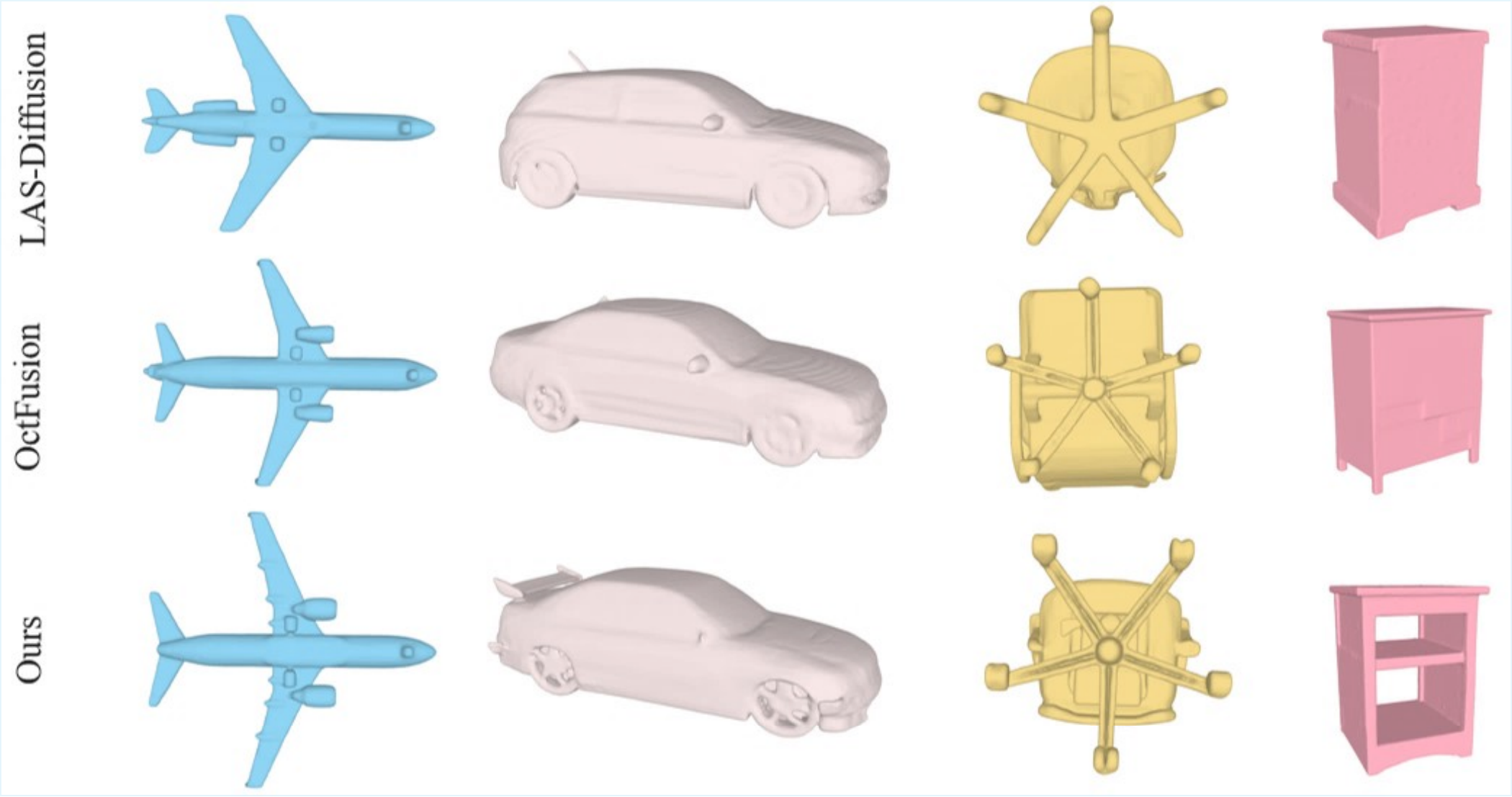} % Reduce the figure size so that it is slightly narrower than the column. Don't use precise values for figure width.This setup will avoid overfull boxes.
\caption{Details comparison of HierOctFusion with OctFusion and LAS-Diffusion. Our method is capable of generating more refined local structures, such as engines and wheels.}
\label{fig4}
\end{figure}

Figure \ref{fig4} shows a direct comparison of locally complex structures generated by our method and other methods. In the airplane examples, our model captures the subtle curvature and mounting details of underside engines, including intake shapes and exhaust contours. For cars, our approach reproduces the full tread pattern on each tire. These visual outcomes demonstrate that incorporating semantic part information through our cross-attention module gives the diffusion model a stronger understanding of fine geometric structures, leading to more accurate and detailed local features and an overall improvement in 3D shape quality.

\begin{figure}[h]
\centering
\includegraphics[width=0.9\columnwidth]{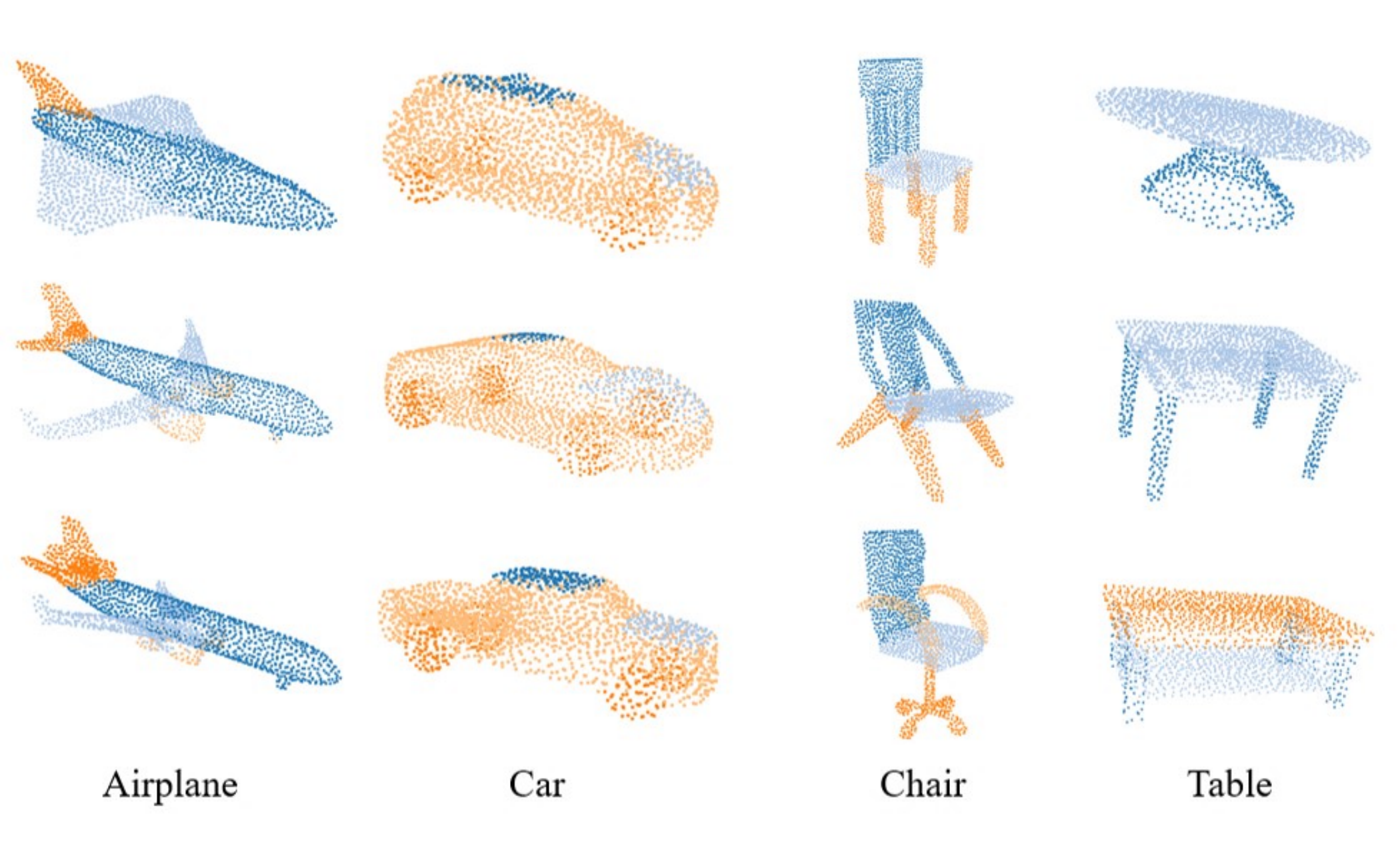} % Reduce the figure size so that it is slightly narrower than the column. Don't use precise values for figure width.This setup will avoid overfull boxes.
\caption{Segmentation results on four categories (airplane, car, chair and table) from the ShapeNet dataset. Airplanes are segmented into wings, fuselages, empennages and engines; cars are segmented into roofs, hoods, bodies and tires; chairs are segmented into backrests, seats, legs and armrests; while tables are segmented into tabletops, legs and drawers.}
\label{fig5}
\end{figure}

Our pre–trained DGCNN reliably decomposes ShapeNet objects into semantic regions, using four parts for airplanes, cars and chairs and three parts for tables, even when individual instances have missing components or unclear boundaries. In most cases the network assigns the vast majority of points to their correct parts, producing robust part-level feature descriptors. These descriptors are then fed into our diffusion framework to improve its ability to learn category-specific shape priors. Representative segmentation outputs are shown in Figure \ref{fig5}, demonstrating how this decomposition isolates local structures for subsequent generation.

\begin{figure}[t]
\centering
\includegraphics[width=0.9\columnwidth]{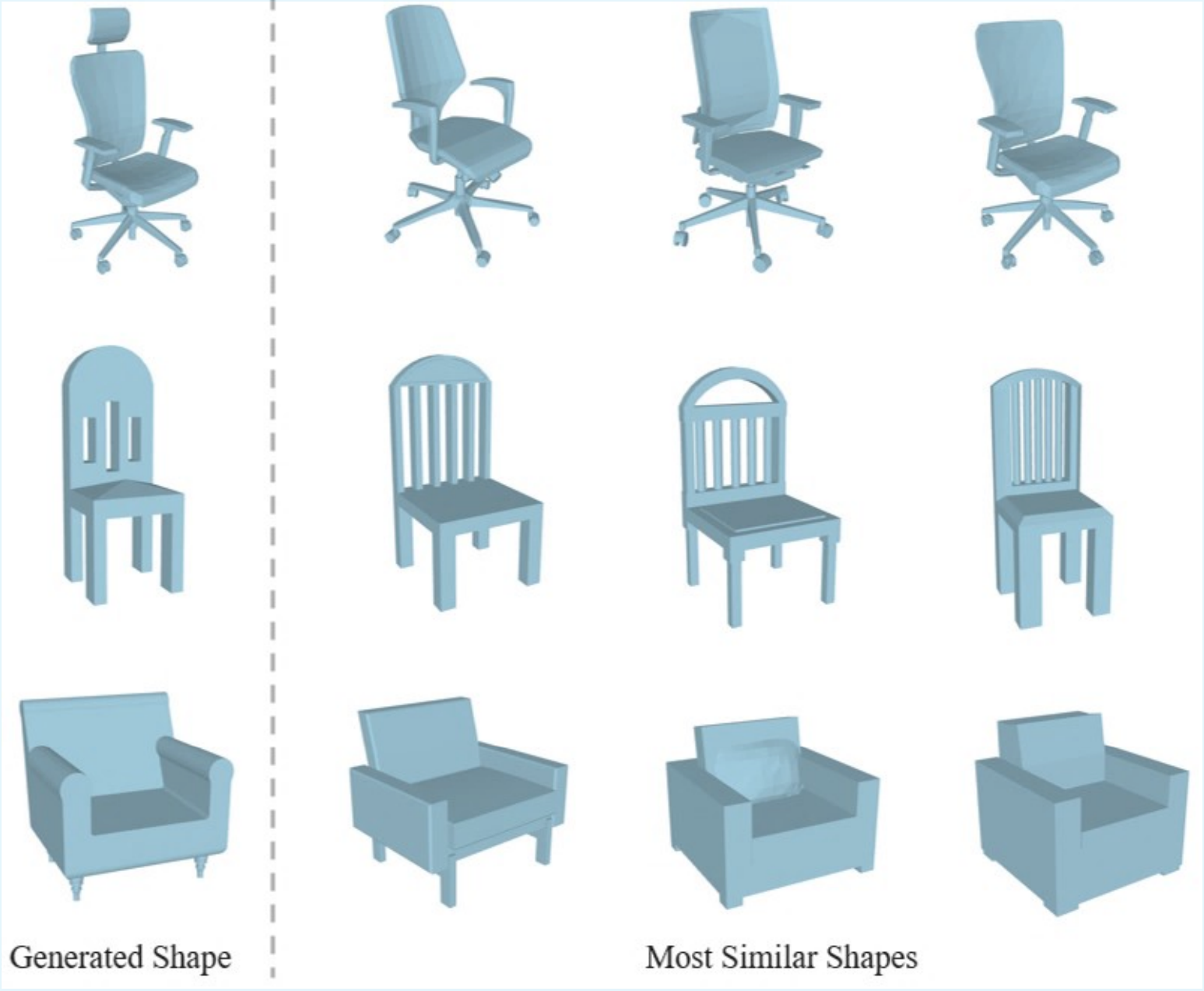} % Reduce the figure size so that it is slightly narrower than the column. Don't use precise values for figure width.This setup will avoid overfull boxes.
\caption{The generated shapes of HierOctFusion (Left) and the three nearest shapes in ShapeNet-Seg (right) retrieved from the training dataset according to their Chamfer distances.}
\label{fig6}
\end{figure}

Figure \ref{fig6} presents some generated samples of HierOctFusion with the three most similar shapes in ShapeNet-Seg. The results demonstrate that our method generates objects with significant diversity and novelty, effectively avoiding the production of repetitive outputs or overfitting to the training data distribution.

\subsection{Ablation Study}
In this section, we analyze the impact of several key components of our proposed method, primarily focusing on the multi-scale diffusion model framework, the proposed part-whole-hierarchy message passing module, and the efficiency comparison with previous methods. For the experiments, we choose category "chair" as the dataset and shading-image-based FID as the evaluation metric.

\subsubsection{Investigation on the proposed modules.}

\begin{table}[h]
\centering
\begin{tabular}{c ccc}
\toprule
Method & Dataset & Category & FID\\
\midrule
Baseline & ShapeNet-Seg & chair & 16.15\\
$\mathcal{F}_1^*+\mathcal{F}_2$ & ShapeNet-Seg & chair & 16.14\\
$\mathcal{F}_1+\mathcal{F}_2^*$ & ShapeNet-Seg & chair & 16.10\\
$\mathcal{F}_1^*+\mathcal{F}_2^*$ & ShapeNet-Seg & chair & \textbf{16.07}\\
\bottomrule
\end{tabular}
\caption{Ablation studies on the effects of adding the part-whole-hierarchy message passing module at different scale. $\mathcal{F}_1$ denotes the coarse-scale, $\mathcal{F}_2$ represents the fine-scale, and the subscript * indicates adding the proposed module.}
\label{tab4}
\end{table}

Table \ref{tab4} presents the influence of our message passing module on different scales. The results reveal that adding the module solely in the coarse-scale yields marginal improvement, whereas its inclusion in the fine-scale plays a dominant role in quality enhancement. We attribute this phenomenon to the distinct roles of each scale: the coarse-scale primarily establishes the global framework and basic geometric structure. Consequently, the module provides limited benefits at this scale. In contrast, the fine-scale specializes in fine-grained optimization and the model effectively leverages part features to guide the optimization of complex local regions, where the geometric priors significantly improve generation quality. Experimental results confirm that our part-whole-hierarchy message passing mechanism exhibits strong compatibility with the multi-scale diffusion framework, particularly in the fine-scale where part features can fully exert their guidance.

\subsubsection{Evaluate the efficiency.}

\begin{table}[h]
\centering
\resizebox{.95\columnwidth}{!}{
\begin{tabular}{c cccc}
\toprule
Method & Node Number* & Node Number** & Memory & Inference Time\\
\midrule
LAS-Diffusion & 262,144 & 124,156 & 1.06G & 66.1ms\\
XCube & 4096 & 74,761 & 12.76G & 135.3ms\\
OctFusion & 4096 & 11,634 & 0.69G & 48.2ms\\
\midrule
HierOctFusion & 4096 & 11,634 & 0.83G & 54.5ms\\
\bottomrule
\end{tabular}
}
\caption{Efficiency comparison with prior works. The average octree node number, the GPU memory and the inference time on a NVIDIA 3090 GPU with batch size 1 are reported. The subscript * and ** denote the coarse-scale and the fine-scale of corresponding methods.}
\label{tab5}
\end{table}

As demonstrated in Table \ref{tab5}, our method maintains computational efficiency comparable to prior approaches. The results indicate that our framework preserves the same node count as OctFusion across both scales. Despite the incorporation of additional modules, our lightweight design ensures minimal impact on both memory and inference time. Consequently, our approach achieves optimal efficiency among all comparative methods except OctFusion. These findings validate the high efficiency of our module design, which introduces negligible computational overhead while simultaneously improving generation quality and maintaining superior efficiency.

\vspace{-0.05in}
\section{Conclusion}
\vspace{-0.05in}
This paper introduces HierOctFusion, a multi-scale octree-based diffusion model for 3D object generation that implements part-level generation through part-whole-hierarchy message passing. The model obtains detailed part information by leveraging a pre-trained part segmentation model and fuses these part features with octree representations via a lightweight module based on cross-attention. By embedding this module at multiple depths within the octree diffusion framework, our approach enables hierarchical transmission of local geometric details encoded by the part features. Experimental results on single-object generation benchmarks confirm that HierOctFusion achieves both high fidelity and computational efficiency. In future work, we will further optimize the method and explore its extension to full scene generation and interactive 3D object editing tasks.

\bibliography{aaai2026}

\end{document}